\documentclass[conference]{IEEEtran}
\IEEEoverridecommandlockouts

\usepackage{cite}
\usepackage{amsmath,amssymb,amsfonts}
\usepackage{algorithmic}
\usepackage{graphicx}
\usepackage{textcomp}
\usepackage{xcolor}
\usepackage[caption=false,font=footnotesize]{subfig}
\usepackage{booktabs}
\usepackage{stfloats}

\def\BibTeX{{\rm B\kern-.05em{\sc i\kern-.025em b}\kern-.08em
    T\kern-.1667em\lower.7ex\hbox{E}\kern-.125emX}}

\begin{document}

\title{Prediction-Informed Landing Window Planning for UAV Landing on USV in High Sea States}

\author{%
\begin{minipage}[t]{0.24\textwidth}\centering
{\normalsize Kaizhen Li}\par\vspace{4pt}
{\itshape School of Mechatronic Engineering and Automation}\par
{\itshape Shanghai University}\par
Shanghai, China\par
likaizhen@shu.edu.cn
\end{minipage}\hfill
\begin{minipage}[t]{0.24\textwidth}\centering
{\normalsize Xi Zhou}\par\vspace{4pt}
{\itshape School of Mechatronic Engineering and Automation}\par
{\itshape Shanghai University}\par
Shanghai, China\par
xizhoushu@shu.edu.cn
\end{minipage}\hfill
\begin{minipage}[t]{0.24\textwidth}\centering
{\normalsize Dan Zhang}\par\vspace{4pt}
{\itshape School of Mechatronic Engineering and Automation}\par
{\itshape Shanghai University}\par
Shanghai, China\par
dan.zhang@shu.edu.cn
\end{minipage}\hfill
\begin{minipage}[t]{0.24\textwidth}\centering
{\normalsize Xiaowei Li*}\par\vspace{4pt}
{\itshape School of Mechatronic Engineering and Automation}\par
{\itshape Shanghai University}\par
Shanghai, China\par
xwli@staff.shu.edu.cn
\end{minipage}}

\maketitle

\begin{figure*}[!b]
\centering
\includegraphics[width=\textwidth]{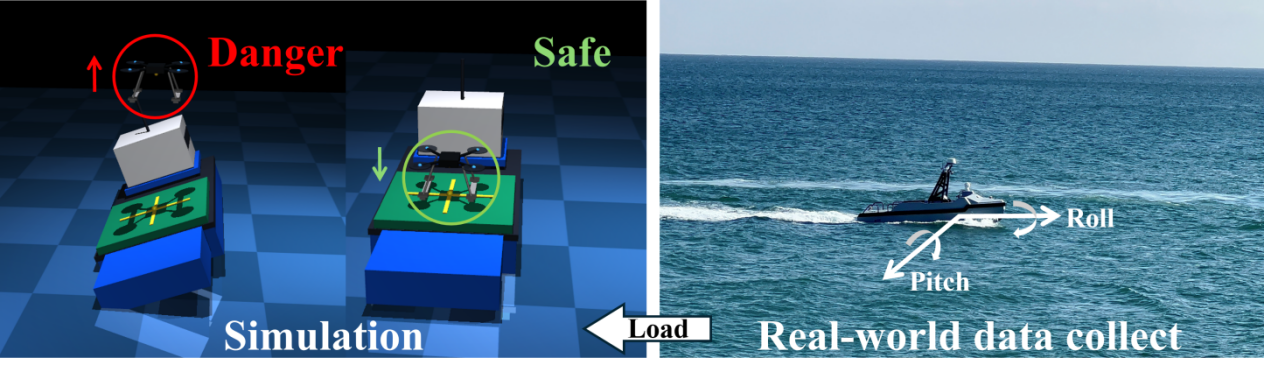}
\caption{Overview of the prediction-informed landing framework. Left: MuJoCo simulation driven by real sea-state USV motion data, where red and green marks indicate dangerous and safe landing windows. Right: the real USV from which roll and pitch data are collected and loaded into the simulation.}
\label{fig:overview}
\end{figure*}

\begin{abstract}
Autonomous landing of unmanned aerial vehicle (UAV) on unmanned surface vehicle (USV) is difficult in high sea states. Existing hierarchical landing methods often infer future USV motion only implicitly from short observation histories, which limits their ability to adapt to severe and rapidly changing sea conditions. This paper presents a prediction-informed hierarchical control framework. A pretrained model is developed to explicitly predict the USV attitude during high-level landing decisions. The prediction helps identify suitable landing windows and choose the right touchdown time, while a conventional low-level flight controller maintains UAV stability and tracks motion commands. To evaluate the framework under realistic disturbances, a real-sim validation setup is constructed in which USV-motion data collected in real sea conditions are transferred to a MuJoCo USV. The experiments demonstrate the feasibility of using explicit attitude prediction to improve landing-window selection and support timing-aware UAV landing planning in rapidly varying marine environments.
\end{abstract}


\section{Introduction}
Unmanned aerial vehicle (UAV) and unmanned surface vehicle (USV) are increasingly used as a complementary robotic team in search and rescue, offshore inspection, environmental monitoring, and maritime surveillance. UAV provide rapid aerial observation and wide-area coverage, whereas USV provide endurance and direct access to the water surface. Reliable autonomous landing is therefore essential for sustained UAV-USV operation because it enables UAV recovery, recharging, and redeployment without continuous human intervention.

Landing a UAV on a moving USV is particularly challenging in high sea states \cite{gupta2023landing}. Wave-induced heave, roll, and pitch continuously change the relative pose and velocity between the vehicle and the deck. During the final descent, USV attitude can vary substantially within the time required for the UAV to reach the deck, so a touchdown condition that is safe at the current instant may become unfavorable before contact. Wind, sensing uncertainty, and control delay further reduce the available touchdown margin. The landing problem therefore involves not only stabilizing the vehicle and tracking the USV, but also choosing a touchdown time that is consistent with the future motion of the deck.

This temporal decision becomes more difficult as sea conditions become more dynamic. A controller that reacts only to the current USV attitude may begin to descend when the platform is temporarily favorable but fail to anticipate the rapid closure of the corresponding landing window. Conversely, an overly conservative controller may miss short favorable intervals and increase the recovery time. These observations motivate a landing architecture that preserves reliable low-level flight control while providing the high-level decision layer with explicit information about future platform attitude.

In this paper, we develop a prediction-informed hierarchical control framework for UAV landing on USV in high sea states. A pretrained temporal model forecasts the short-term evolution of USV attitude, and the predicted attitude is used to evaluate feasible touchdown windows and guide the high-level landing decision. A conventional low-level flight controller remains responsible for vehicle stabilization and motion tracking. The framework is evaluated in a real-to-simulation setting, where USV-motion data collected under real sea conditions are transferred to a MuJoCo landing platform.

The main contributions of this paper are as follows:
\begin{itemize}
\item A prediction-informed hierarchical landing framework is developed by coupling explicit platform-attitude forecasting with high-level landing-window planning and conventional low-level flight control.
\item A pretrained temporal prediction model is introduced to provide future attitude information for touchdown-window assessment under rapidly varying sea states.
\item A real-to-simulation validation setup is established by replaying platform-motion data collected in real sea conditions on a MuJoCo platform, enabling evaluation under realistic disturbance patterns.
\end{itemize}

\section{Related Work}
Learning-based methods have been investigated to improve the adaptability of UAV landing under uncertain platform motion. Compared with controllers that depend on an accurate analytical model, such as NMPC-based cooperative UAV-USV tracking and landing \cite{li2023nmpc} and synchronized-motion-based cooperative landing \cite{li2022synchronized}, reinforcement learning (RL) can learn a feedback policy from repeated interaction with a simulator or physical system. Polvara et al. explored end-to-end UAV landing through deep reinforcement learning, mapping visual observations to landing actions \cite{polvara2018toward}. Rodriguez-Ramos et al. extended deep RL to landing on moving platforms, demonstrating that a learned policy can adapt its control behavior to platform motion \cite{rodriguez2019deep}. Subsequent work considered autonomous multi-rotor landing and vision-based ship landing under disturbance, further showing the potential of learning-based policies in dynamic landing scenarios \cite{goldschmid2024reinforcement}, \cite{saj2022robust}.

Despite this adaptability, fully end-to-end learning is difficult to deploy in safety-critical flight systems. High-dimensional visual inputs, the simultaneous learning of stabilization and landing decisions, and the absence of an explicit temporal representation can increase training complexity and reduce interpretability. These limitations have motivated hierarchical formulations, in which a learned policy makes high-level landing decisions while a conventional flight controller maintains attitude stability and tracks motion commands. WaveLander is representative of this direction: its recurrent RL policy uses a compact platform-relative observation and outputs a scalar vertical-velocity reference, while the low-level controller handles stabilization and tracking \cite{li2026wavelander}.

The main limitation of current hierarchical learning-based landing methods is their limited awareness of future landing windows. In WaveLander, timing-aware behavior is learned from short observation histories, recurrent hidden states, and attitude-aware reward shaping. The policy can therefore slow down, hold, or retreat when the current platform condition is unfavorable, but its prediction of future platform motion remains implicit. Under rapidly changing high-sea-state conditions, this reactive mechanism may identify an unfavorable touchdown condition only after a safe landing interval has started to close. The policy is consequently required to infer both the platform trend and the appropriate touchdown timing from limited short-term evidence.

Explicit platform-motion prediction provides a complementary solution to this limitation. Li et al. used recurrent prediction to estimate USV attitude and exploit favorable motion phases during cooperative landing \cite{li2022synchronized}. Such methods demonstrate that future attitude information can support temporal coordination, but the prediction module is commonly coupled to a specific synchronization or tracking controller. The predicted motion is not necessarily exposed as a general landing-window interface for a hierarchical planner. In contrast, the present work treats attitude prediction as an independent high-level information source: the predicted attitude evolution is used to assess candidate touchdown windows and guide the descent decision, while the conventional low-level flight-control stack is retained. This formulation separates prediction, landing-window planning, and stabilization, and is intended to improve temporal awareness without requiring the learned component to directly control UAV actuators.

\section{Methodology}

\subsection{System Overview}

\begin{figure}[!t]
\centering
\includegraphics[width=\linewidth]{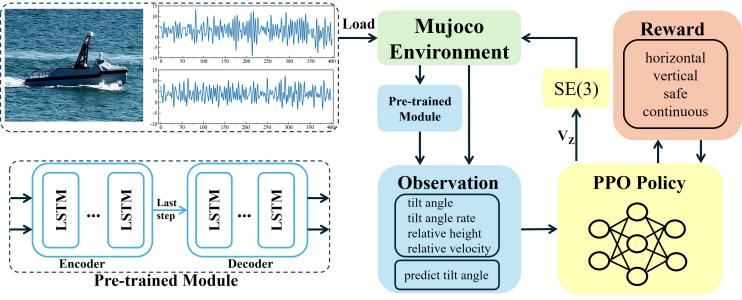}
\caption{Prediction-informed landing architecture.}
\label{fig:architecture}
\end{figure}

As shown in Fig.~\ref{fig:architecture}, the system adopts a prediction-informed hierarchical control architecture. USV motion data collected under real sea conditions are used to pretrain a model that predicts short-term future attitude changes. The motion is then loaded into the MuJoCo environment to drive the dynamic platform. The current deck tilt, tilt rate, UAV relative height, relative vertical velocity, and predicted attitude are combined to form the observation input to the PPO policy. PPO outputs a continuous vertical-velocity command, while the low-level $SE(3)$ controller allocates the motor thrusts. The reward function jointly considers horizontal alignment, descent progress, touchdown safety, and platform attitude, guiding the UAV to select an appropriate landing time.

\subsection{Pretrained Module}
The prediction module is an encoder-decoder LSTM trained before RL optimization. It uses real USV motion data sampled at 0.1~s. A history of 20 samples is used to predict the next 10 samples, providing a 1.0~s forecast:
\begin{equation}
\hat{\mathrm{roll}}, \hat{\mathrm{pitch}} = F_{\mathrm{LSTM}}\left(\mathrm{roll}, \mathrm{pitch}, w_{1}, w_{2}\right)
\label{eq:lstm}
\end{equation}
\begin{equation}
\hat{\phi}_{t} = \sqrt{\hat{\mathrm{roll}}^{2} + \hat{\mathrm{pitch}}^{2}}
\label{eq:tilt}
\end{equation}
The LSTM has 256 hidden units and a linear output layer. Each input window is normalized independently and restored after prediction. The model is trained with MSE loss, while MAE is used for monitoring. Adam, weight decay, learning-rate scheduling, and early stopping are adopted. During online inference, the forecast is updated every 0.1~s and sent to the RL environment.

\subsection{Prediction-Informed Model}
The observation combines the current landing state with the predicted deck motion:
\begin{equation}
o_{t} = \left[ h_{t}, \dot{h}_{t}, \theta_{t}, \dot{\theta}_{t}, \hat{\phi}_{t+1}, \ldots, \hat{\phi}_{t+10} \right]^{T} \in \mathbb{R}^{14}
\label{eq:observation}
\end{equation}
The first four elements describe the current relative height, relative vertical velocity, platform tilt, and tilt rate. The remaining ten elements provide the predicted roll sequence. The PPO policy maps this observation to a continuous relative vertical-velocity command:
\begin{equation}
a_{t} = \pi\left(o_{t}\right) = v_{z}, \quad v_{z} \in [-0.8, 0.8]~\mathrm{m \cdot s^{-1}}
\label{eq:action}
\end{equation}
The predicted sequence improves awareness of future landing windows and contributes to reward evaluation, but it does not directly replace the policy action.

As the high-level decision layer, the PPO policy outputs only the vertical velocity reference. The $SE(3)$ controller then computes the desired acceleration and thrust direction, generates the desired attitude and moments, and allocates the resulting wrench to the four motors.

\subsection{Reward Design}
The reward is implemented as an additive multi-component function:
\begin{equation}
R_{t} = R_{\mathrm{prog}} + R_{\mathrm{td}} + R_{\mathrm{forecast}} + R_{\mathrm{rate}} + R_{\mathrm{safety}}
\label{eq:reward}
\end{equation}
The progress term tracks a height-dependent target descent velocity. The target is faster at high altitude and gradually approaches zero near the contact height:
\begin{equation}
R_{\mathrm{prog}} = -\left(\dot{h}_{t} - v_{\mathrm{des}}\left(h_{t}\right)\right)^{2}
\label{eq:progress}
\end{equation}
The touchdown term is activated when $h_{t} \le 0.50$~m and $e_{xy} \le 0.25$~m. Otherwise, it is zero:
\begin{equation}
R_{\mathrm{td}} = 1000\, e^{-\frac{\dot{h}_{t}^{2}}{0.405}}\, e^{-\frac{\theta_{t}^{2}}{0.01}}
\label{eq:touchdown}
\end{equation}
The forecast term includes forecast stability $S_{t}$, and peak-tilt excess $E_{t}$, which are computed from the predicted attitude sequence $\hat{\phi}_{t}$:
\begin{equation}
S_{t} = e^{-\frac{\mathrm{mean}\left(\hat{\phi}_{t}^{2}\right)}{0.405}}
\label{eq:stability}
\end{equation}
\begin{equation}
E_{t} = \mathrm{clip}\left(\frac{\max_{i}\left|\hat{\phi}_{t}\right| - \theta_{\mathrm{safe}}}{\theta_{\mathrm{limit}} - \theta_{\mathrm{safe}}}, 0, 1\right)
\label{eq:excess}
\end{equation}
\begin{equation}
R_{\mathrm{forecast}} = 1.5\, S_{t} - 8\, E_{t}
\label{eq:forecast}
\end{equation}
The action-rate term penalizes abrupt changes in the vertical-velocity command:
\begin{equation}
R_{\mathrm{rate}} = -10\left(v_{z}(t) - v_{z}(t-1)\right)^{2}
\label{eq:rate}
\end{equation}
The safety term adds penalties when the tilt exceeds 15 degrees, the descent speed is below $-1.0$~m/s. At episode termination or truncation, the agent receives $+5000$ for a successful landing and $-2000$ otherwise.

\subsection{Training Setup}
The pretrained predictor is trained using high-sea-state USV motion data. The dataset is divided into training and test subsets in a 7:3 ratio. A sliding-window strategy is used to preserve the temporal structure of the motion sequence. Each window contains 20 historical samples as input and the following 10 samples as the prediction target. With a sampling interval of 0.1~s, the model uses a 2.0~s history to forecast the platform motion over the next 1.0~s.

The policy is trained in a MuJoCo environment driven by recorded or generated USV motion data. Each episode starts with the UAV at an altitude of 2.0~m and a random horizontal position within $[-2, 2]$~m. The observation includes the relative height, relative vertical velocity, deck tilt, deck tilt rate, and a 10-step roll forecast. The action is a continuous vertical-velocity command in $[-0.8, 0.8]$~m/s.

PPO is used for policy optimization. The actor and critic each contain two hidden layers with 128 units. The learning rate is $3 \times 10^{-4}$, the rollout length is 2048 steps, the batch size is 64, and each rollout is updated for 10 epochs. The discount factor, GAE parameter, and clipping range are $\gamma = 0.99$, $\lambda = 0.95$, and 0.2, respectively. The experiments are conducted using an NVIDIA GeForce RTX 5060 Ti GPU.

\begin{figure*}[!t]
\centering
\begin{minipage}[t]{0.49\textwidth}
  \centering
  \includegraphics[width=\linewidth]{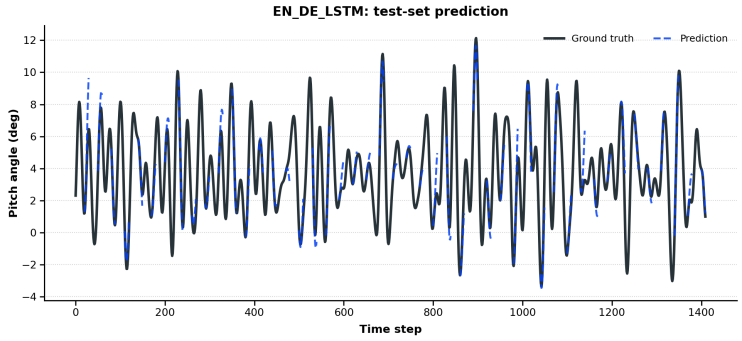}
  \caption{Test-set prediction of the USV pitch angle.}
  \label{fig:pitch}
\end{minipage}
\hfill
\begin{minipage}[t]{0.49\textwidth}
  \centering
  \includegraphics[width=\linewidth]{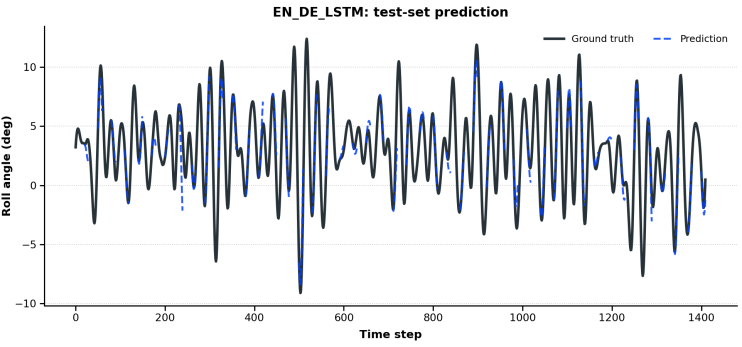}
  \caption{Test-set prediction of the USV roll angle.}
  \label{fig:roll}
\end{minipage}
\end{figure*}

\section{Experiments and Analysis}

\subsection{Pretrained Module Behavior Analysis}
The predictor uses the USV roll and pitch angles as a two-channel input sequence. An LSTM encoder processes the 20 historical samples and compresses their temporal information into the final hidden and cell states. The decoder is initialized with this final encoder state and the last observed attitude, and directly produces the complete 10-step prediction sequence.

As shown in Fig.~\ref{fig:pitch} and Fig.~\ref{fig:roll}, the predicted curves closely follow the measured platform motion over the 1.0~s horizon, accurately capturing the temporal evolution and peak values of both roll and pitch. Small amplitude differences and phase deviations may occur near rapid peaks, but the overall prediction is sufficiently accurate to provide the landing policy with short-term information about upcoming platform motion and potential landing windows.

Prediction performance is quantitatively evaluated using the root mean square error (RMSE) and mean absolute error (MAE). Two feature extraction architectures, LSTM and Informer, are compared for predicting the USV roll and pitch. As reported in Tables~\ref{table:pitch} and \ref{table:roll}, the LSTM-based model achieves the lowest RMSE and MAE values for both prediction tasks, demonstrating superior accuracy and effectiveness in capturing the underlying temporal patterns of the data.

\begin{table}[!t]
\caption{Metric of Predicted USV Pitch Angle}
\label{table:pitch}
\centering
\begin{tabular}{ccc}
\toprule
\textbf{Model} & \multicolumn{2}{c}{\textbf{Metric}} \\
\cmidrule(lr){2-3}
 & \textbf{RMSE} & \textbf{MAE} \\
\midrule
Informer & 1.286 & 0.841 \\
LSTM & 0.913 & 0.557 \\
\bottomrule
\end{tabular}

\vspace{2ex}

\caption{Metric of Predicted USV Roll Angle}
\label{table:roll}
\centering
\begin{tabular}{ccc}
\toprule
\textbf{Model} & \multicolumn{2}{c}{\textbf{Metric}} \\
\cmidrule(lr){2-3}
 & \textbf{RMSE} & \textbf{MAE} \\
\midrule
Informer & 1.274 & 0.833 \\
LSTM & 0.997 & 0.617 \\
\bottomrule
\end{tabular}
\end{table}

\subsection{Policy Behavior Analysis}

\begin{figure*}[!t]
\centering
\subfloat[]{\includegraphics[width=0.48\textwidth]{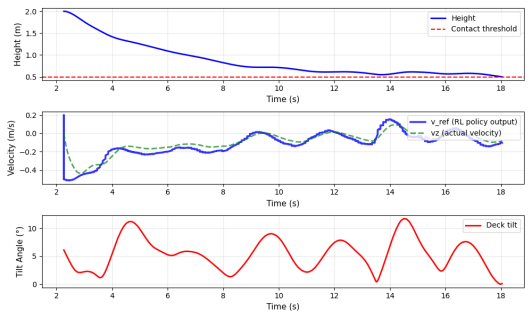}%
\label{fig:seg_a}}
\hfil
\subfloat[]{\includegraphics[width=0.48\textwidth]{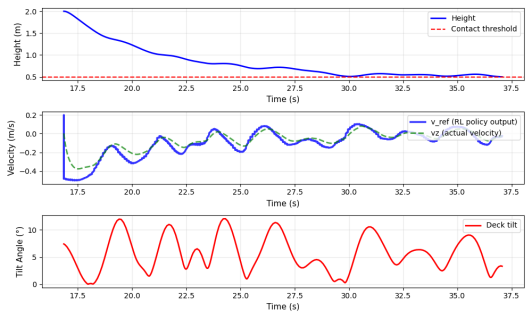}%
\label{fig:seg_b}}

\subfloat[]{\includegraphics[width=0.48\textwidth]{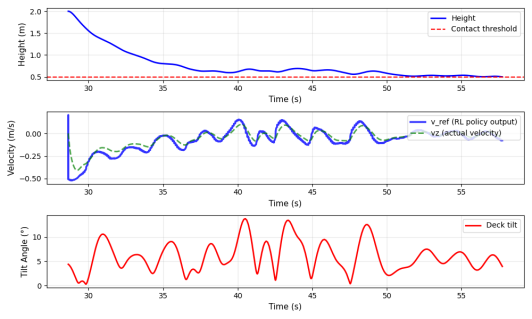}%
\label{fig:seg_c}}
\hfil
\subfloat[]{\includegraphics[width=0.48\textwidth]{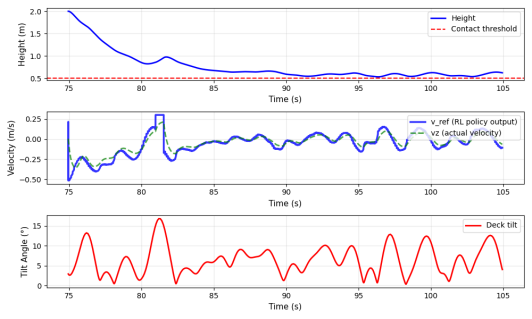}%
\label{fig:seg_d}}
\caption{Motion segments of the landing policy. Each segment shows the UAV height, the PPO vertical-velocity command and the actual vertical velocity, and the USV deck tilt. The dashed lines mark the contact or safety thresholds.}
\label{fig:policy}
\end{figure*}

Fig.~\ref{fig:policy} contains four motion segments. Each segment shows the UAV height, the PPO vertical-velocity command and the actual vertical velocity, the USV deck tilt. The dashed lines mark the contact or safety thresholds.

PPO uses the current landing state and the predicted deck attitude to adjust the vertical velocity. It descends when the platform is favorable and slows, hovers, or briefly ascends when a large attitude change is predicted.

For example, in Fig.~\ref{fig:seg_a}, the deck tilt increases rapidly at about 5~s. The policy changes the command from descent to upward at 4~s, allowing the UAV to avoid entering the unfavorable motion window. Around 17--18~s, the tilt is small and the UAV maintains a near-zero vertical velocity before approaching the contact threshold. This shows that prediction enables the UAV to adjust the landing timing in advance.

Overall, the UAV vertical-velocity command leads the USV attitude by several phases. When the deck angle is about to increase rapidly, the policy reduces the descent rate in advance or generates an opposite-direction velocity command, allowing the UAV to slow down, hover, or climb before entering the unfavorable landing window.

\section{Conclusion}
This paper presented a prediction-informed hierarchical control framework for UAV landing on a USV under high sea states. A pretrained model explicitly forecasts the roll and pitch motion of the USV, and the predicted attitude is incorporated into the PPO observation and reward design. The high-level PPO policy generates a vertical-velocity reference, and the low-level $SE(3)$ controller tracks this reference while maintaining the stability of the UAV.

The real-sim results show that explicit attitude prediction improves the policy's awareness of future landing windows and is more suitable for rapidly varying sea conditions. The UAV can reduce its descent rate, hover, or generate an opposite-direction velocity command before the USV attitude changes significantly. This behavior allows the policy to adjust the landing timing continuously without relying on manually designed rules to switch between descending, hovering, and ascending states.

However, the current work still has limitations. Owing to the limited data collected under high sea states, only the pitch and roll of the USV are considered in the prediction and planning framework. Future work will further investigate the influence of other motion variables, such as heave and displacement, on the landing-window selection. Moreover, the proposed policy will be transferred to a physical UAV-USV platform to further validate its effectiveness in real-world experiments.

\section*{Acknowledgment}
The financial support from the National Natural Science Foundation of China (No.~62225308), and Shanghai Science and Technology Innovation Project (No.~XTCX-KJB002-2025-2-03) is also acknowledged. (*Corresponding author: Xiaowei Li.)

\end{document}